\documentclass[9pt, conference]{IEEEtran}
\IEEEoverridecommandlockouts
\usepackage{cite}
\usepackage{amsmath,amssymb,amsfonts}
\usepackage{graphicx}
\usepackage{textcomp}
\usepackage{booktabs}
\usepackage{xcolor}
\usepackage{algorithm} 
\usepackage{algpseudocode} 
\usepackage{hyperref}

\begin{document}
\title{Delayed Fusion: Integrating Large Language Models into First-Pass Decoding in End-to-end Speech Recognition}
\author{
\IEEEauthorblockN{\textit{\shortstack{Takaaki Hori$^1$, Martin Kocour$^{2*}$\thanks{*This work was done during an internship at Apple.}, Adnan Haider$^1$, Erik McDermott$^1$, Xiaodan Zhuang$^1$}\vspace{.7\baselineskip}}}        
\IEEEauthorblockA{{\hspace{0.5cm}$^1$Apple, \hspace{0.5cm} $^2$Brno University of Technology}}
}
\maketitle
\begin{abstract}
This paper presents an efficient decoding approach for end-to-end automatic speech recognition (E2E-ASR) with large language models (LLMs). 
Although shallow fusion is the most common approach to incorporate language models into E2E-ASR decoding, we face two practical problems with LLMs. (1) LLM inference is computationally costly. (2) There may be a vocabulary mismatch between the ASR model and the LLM.
To resolve this mismatch, we need to retrain the ASR model and/or the LLM, which is at best time-consuming and in many cases not feasible.
We propose \textit{delayed fusion}, which applies LLM scores to ASR hypotheses with a delay during decoding and enables easier use of pre-trained LLMs in ASR tasks. This method can reduce not only the number of hypotheses scored by the LLM but also the number of LLM inference calls. It also allows re-tokenizion of ASR hypotheses during decoding if ASR and LLM employ different tokenizations.
We demonstrate that delayed fusion provides improved decoding speed and accuracy compared to shallow fusion and N-best rescoring using the LibriHeavy ASR corpus and three public LLMs, OpenLLaMA 3B \& 7B and Mistral 7B. 
\end{abstract}
\begin{IEEEkeywords}
speech recognition, large language model, decoding, delayed fusion
\end{IEEEkeywords}
\section{Introduction}
\label{sec:intro}
Large language models (LLMs) have shown their tremendous power of language understanding and generation in various domains~\cite{achiam2023gpt,touvron2023llama,chowdhery2023palm,jiang2023mistral}.
LLMs, including many publicly available ones ~\cite{minaee2024large}, are typically Transformer models \cite{vaswani2017attention} with billions of parameters trained on vast amounts of text data.
Towards effectively exploiting LLMs for ASR, researchers are very actively exploring various LLM-based ASR models and decoding approaches~\cite{rubenstein2023audiopalm,wu2023decoder,mittal2024salsa}.

Most ASR systems employ an external language model to improve recognition accuracy.
If similarly applying an LLM using conventional shallow fusion, we face two practical problems.
(1) LLM inference is computationally demanding, making it costly to directly apply shallow fusion during beam search, which requires many LLM inference calls, especially in frame-synchronous decoding.
(2) There may be a vocabulary mismatch between ASR model and LLM.
LLMs typically have a much larger vocabulary compared to end-to-end ASR models. To apply shallow fusion, ASR model and LLM need to have identical vocabularies. To match one vocabulary to another, either the ASR model or the LLM needs to be retrained. However, training an ASR model on the LLM vocabulary leads to an out-of-vocabulary problem, since the paired data used for training ASR models is limited and does not cover LLM vocabulary sufficiently well. On the other hand, training an LLM on the ASR vocabulary is expensive, time consuming and in many cases infeasible. Furthermore, publicly available pre-trained LLMs cannot be easily adapted to different vocabularies.

$N$-best rescoring is a possible solution to relieve the above problems: first-pass decoding generates $N$-best hypotheses, and then a second-pass rescores the hypotheses using an LLM, after re-tokenization. The rescoring pass is not very expensive if a graphics processing unit (GPU) is available, since rescoring requires only one LLM inference call when first-pass hypotheses are batched. However, $N$ needs to be large enough to make rescoring effective. Especially for long utterances, it is difficult to generate short $N$-best lists that include the correct hypothesis. Increasing the list size imposes a big burden on the first-pass decoding, where more computation and memory are needed for a wider beam, especially when using an auto-regressive E2E model that predicts the next tokens based on all the previous tokens.

We propose \textit{delayed fusion}, where we apply LLM scores during decoding but only after pruning, which dramatically reduces the number of partial hypotheses that need to be scored by the LLM. At the same time, we can wait until a partial hypothesis reaches the end of word to handle different tokenizations between ASR and LLM. Once the decoder detects the end of a word, it re-tokenizes the word, computes the LLM score and adds it to the current partial hypothesis score. This way, LLM scores are incorporated from an early stage of the first-pass decoding, reducing search errors otherwise not recoverable by $N$-best rescoring.

The contributions of this work are:
\begin{itemize}
\item We propose a novel method for efficient LM fusion, which allows us to
    (1) easily compare different LLMs on ASR tasks,
    (2) investigate the effect of prompting LLMs in ASR, and
    (3) use as a baseline system when exploring advanced LLM-based ASR models.
\item  We provide experimental results on ASR accuracy and decoding speed with three public domain LLMs, OpenLLaMA 3B \& 7B~\cite{touvron2023llama2, openlm2023openllama} and Mistral 7B~\cite{jiang2023mistral}, showing that (1) Delayed LLM fusion is fast enough compared to standard neural language model (NLM) fusion, allowing us to obtain improved ASR accuracy from LLMs in the same decoding time, and that (2) Delayed LLM fusion provides significant WER reduction compared to $N$-best rescoring with LLMs.
\end{itemize}

\section{Related work}
\label{sec:related_work}
There are different types of E2E-ASR systems~\cite{prabhavalkar2023end} and many of them employ an external LM to improve recognition accuracy~\cite{hori2017advances,toshniwal2018comparison,Zhou+Zheng+:2022}.
Some LM fusion techniques require retraining of the ASR model to further improve accuracy and adaptability to other domains~\cite{sriram2018cold,stahlberg2018simple,Meng+Parthasarathy+:2020}. 
Unlike such techniques, this paper focuses on improving shallow fusion based decoding, which combines E2E-ASR and LM without retraining or fine-tuning.

In standard shallow fusion decoding, ASR model and LM need to use the same tokenization and vocabulary.
This limitation prevents us from easily applying LLM shallow fusion.
Prior work has investigated delayed LM application using on-the-fly lattice rescoring~\cite{hori2007efficient,sak2010fly} for hybrid ASR and
shallow fusion of a character-based E2E model and a word-based LM for end-to-end ASR, where a space token is used to trigger word-based LM scoring~\cite{hannun2014first,hori2018end}.
However, these approaches do not have a mechanism to control the delay for efficient shallow fusion.
With delayed fusion, we can adjust the timing of LM fusion considering the computation vs. accuracy trade-off as well as tokenization mismatches. Moreover, unlike LLM rescoring \cite{huang2019empirical, xu2022rescorebert, huang2024multilingual,udagawa2022effect} or redecoding \cite{yang2023generative, ma2023n} approaches, delayed fusion can be used in streaming scenarios, similarly to standard shallow fusion.

A recent work related to our approach is SALSA~\cite{mittal2024salsa} since it can handle tokenization mismatches during decoding. However, SALSA integrates pre-trained ASR and LLM by combining their state vectors using additional projection layers, and thus, it needs to train those layers using paired data before decoding. On the other hand, delayed fusion integrates ASR and LLM by combining their scores, and therefore, it does not require any extra layers or training steps.

\section{Method}
\label{sec:method}
\subsection{Delayed fusion concept}
\label{ssec:concept}
Delayed fusion computes LM scores for partial hypotheses during decoding as in shallow fusion. However, such scoring is done after pruning, enabling flexible timing of fusion balancing the trade-off between accuracy and computational cost.  
General auto-regressive decoding with delayed fusion is described in Algorithm \ref{algo:delayed_fusion}.
\begin{algorithm}
	\caption{Auto-regressive decoding with delayed fusion}
    \label{algo:delayed_fusion}
	\begin{algorithmic}[1]
        \State $H_0 \leftarrow \{\texttt{<s>}\}$
        \State $S_{E2E}(\texttt{<s>}) \leftarrow S_{LM}(\texttt{<s>}) \leftarrow 0$
		\For {$t = 1,2,\ldots,T$}
			\State $H_t \leftarrow \textsc{Extend}(H_{t-1})$
            \State $S_{E2E} \leftarrow \textsc{E2EScore}(H_t, S_{E2E})$
            \State $H_t \leftarrow \textsc{Prune}(H_t, S_{E2E}, S_{LM}, K)$
			\If {$\textsc{Fusable}(H_{0:t}, t)=\textsc{True}$},
                \State $S_{LM} \leftarrow \textsc{LMScore}(H_t, S_{LM})$
            \EndIf
		\EndFor
        \State $\hat{H}_T \leftarrow \textsc{Finalize}(H_T)$
        \State $S_{LM} \leftarrow \textsc{LMScore}(\hat{H}_T, S_{LM})$
        \State $\hat{h} \leftarrow \mathop{\mbox{argmax}}_{h\in \hat{H}_T} (S_{E2E}(h) + S_{LM}(h)) $
	\end{algorithmic} 
\end{algorithm}

The method first creates an initial hypothesis list with begin-of-sentence token \texttt{<s>} (line 1) and initializes E2E model score list $S_{E2E}(\texttt{<s>})$ and LM score list $S_{LM}(\texttt{<s>})$ (line 2).
For each step $t$, it extends the previous hypothesis list $H_{t-1}$ to get the current hypothesis list $H_t$, and computes E2E model scores $S_{E2E}(h)$ for $h \in H_t$ (lines 4--5).
Then, it applies pruning for $H_t$ to keep the top $K$ hypotheses based on $S_{E2E}$ and $S_{LM}$ (line 6). If the fusion condition $\textsc{Fusable}(H_{0:t}, t)$ is met,%
it computes LM scores $S_{LM}(h)$ for $h \in H_t$ (lines 7--9).
After $T$ steps, it updates the hypothesis list $H_T$, appending the end-of-sentence token \texttt{</s>} to each $h$ in $H_T$ if necessary (line 11), and also computes LM scores $S_{LM}$ for $H_T$ (line 12).
Finally, it selects the best hypothesis $\hat{h}$ based on $S_{E2E}$ and $S_{LM}$ (line 13).

Algorithm \ref{algo:delayed_fusion} shows the abstract-level decoding steps.
In frame-synchronous decoding, $t$ represents a time frame. For CTC decoding, $S_{E2E}(h)$ must have two elements for alignment paths ending with blank and non-blank labels respectively, which are updated according to the CTC rule~\cite{graves2006connectionist} in $\textsc{E2EScore}(\cdot)$.
In label-synchronous decoding, $t$ represents the number of labels generated for each hypothesis, where all existing hypotheses have the same length. In addition, the algorithm requires a function that decides whether to exit the for loop or not, for example checking whether all existing hypotheses end with \texttt{</s>}.
This step is omitted from the algorithm for simplicity.

Thus, the algorithm is agnostic to the decoding strategy, be it frame-synchronous or label-synchronous decoding, as long as it is auto-regressive. The key step is the LM score computation in line 8, which is performed after pruning.
The timing can be controlled by the function $\textsc{Fusable}(\cdot)$ to reduce the number of LM calls for efficiency.

\subsection{Delayed fusion with LLM}
\label{ssec:delayed_fusion}
\begin{figure}[tbh]
\centering
\centerline{\includegraphics[width=8.8cm]{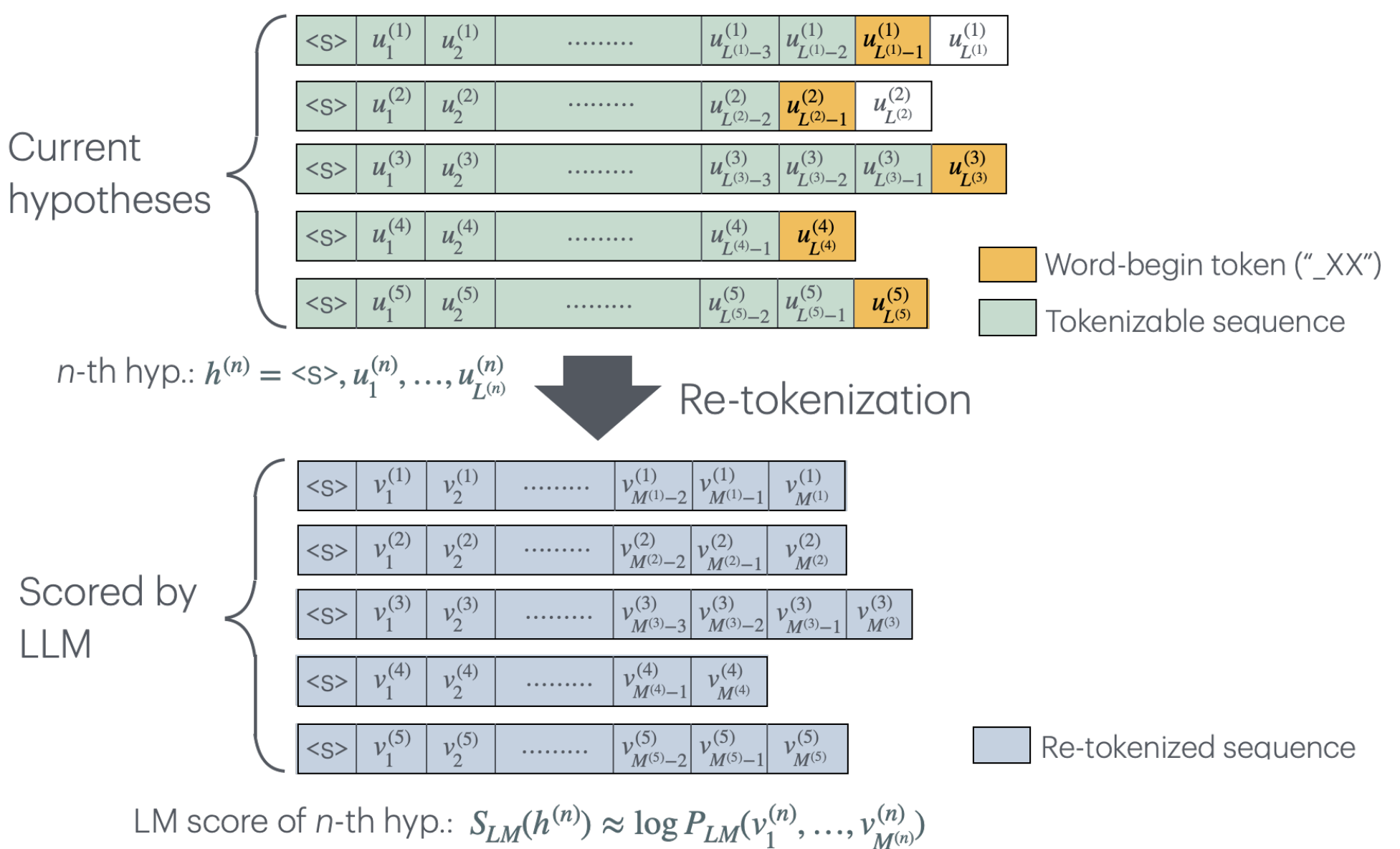}}
\caption{Re-tokenization for delayed LLM fusion.}
\label{fig:retokenization}
\end{figure}

If the ASR model and the LLM were trained with different vocabularies, we need to re-tokenize hypotheses before LLM scoring. However, tokenization may be incorrect for incomplete hypotheses.
Accordingly, we determine a tokenizable sub-sequence as the longest prefix that ends with a word-end token occurring right before a word-begin token, as shown in Fig. \ref{fig:retokenization}. Then, the sequence is re-tokenized using the LLM tokenizer. 
If a standard SentencePiece tokenizer~\cite{kudo2018sentencepiece} is used, each word is tokenized into a unique token sequence. Therefore, the sequence can be re-tokenized into a unique and consistent token sequence, for which an LLM score can be computed correctly.

With delayed fusion, we can call the LLM at any time during decoding. For efficient LLM computation, we propose (1) shortest-hypothesis fusion and (2) fixed-interval fusion. The shortest-hypothesis fusion calls the LLM only when the length of the shortest re-tokenized sequence has increased. The fusion condition is defined as
\begin{align}
\textsc{Fusable}(H_{0:t}, t)=\left\{
\begin{array}{ll}
\textsc{True} & \mbox{if}~\varphi(\bar{H}_{t-1}) < \varphi(\bar{H}_t) \\
\textsc{False} & \mbox{otherwise}
\end{array}
\right.,\nonumber
\end{align}
where $\bar{H}_t$ is a list of re-tokenized sequences obtained from $H_t$ or identical to $H_t$ if there is no tokenization mismatch. $\varphi(\bar{H}_t)$ returns the length of the shortest sequence in $\bar{H}_t$. 
This method allows us to keep the number of LLM calls to at most the number of tokens in the shortest hypothesis.

The fixed-interval fusion calls the LLM at a fixed frame (or label) interval using 
\begin{align}
\textsc{Fusable}(H_{0:t}, t)=\left\{
\begin{array}{ll}
\textsc{True} & \mbox{if}~\bar{H}_{t-I} \ne \bar{H}_t \land t \bmod I=0 \\
\textsc{False} & \mbox{otherwise}
\end{array}
\right.,\nonumber
\end{align}
where $I$ denotes a pre-defined fixed interval.
Increasing $I$ reduces the number of LLM calls. Consequently, a larger $I$ improves the decoding speed but increases the fusion delay, which may cause accuracy degradation.
At the extreme, if $\textsc{Fusable}(\cdot)$ always returns $\textsc{False}$, the algorithm is equivalent to $N$-best rescoring, which in turn can be seen as a special case of delayed fusion.

\begin{figure}[tbh]
\centering
\centerline{\includegraphics[width=8.8cm]{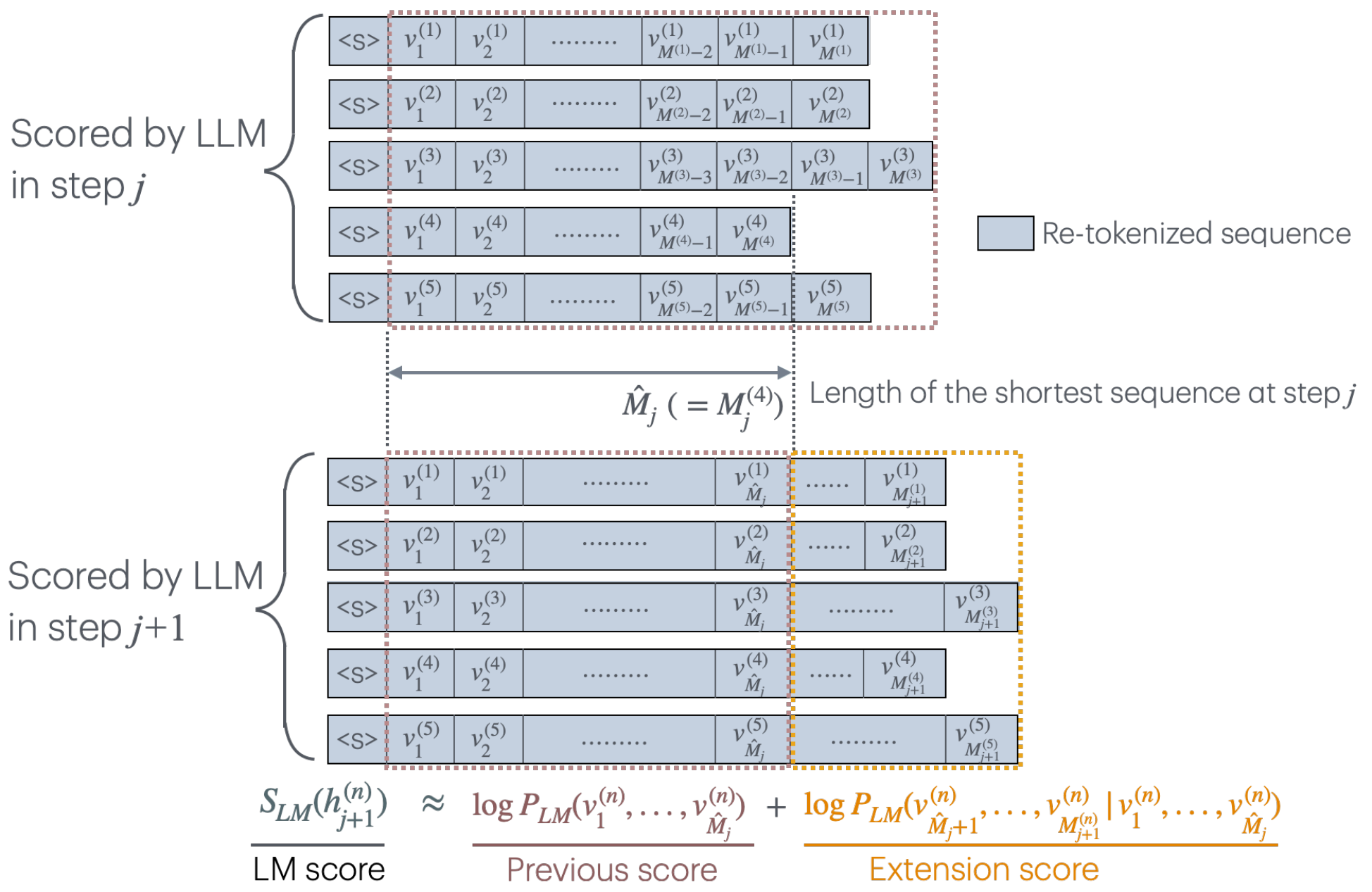}}
\caption{LLM score computation in decoding.}
\label{fig:lm_scoring}
\end{figure}
For efficient LLM fusion, we compute LLM scores for all the current hypotheses at once, where we use hypothesis batching and a key-value cache to take full advantage of GPU acceleration. 
Fig. \ref{fig:lm_scoring} shows how to compute LLM scores at the $(j+1)$-th LLM call.
The LM scores have already been computed for at least $\hat{M}_j$ tokens in the previous step $j$, where $\hat{M}_j$ denotes the length of the shortest sequence at step $j$.
This means that we can use the key-value cache and the LLM scores for the preceding sequence $v^{(n)}_1,\dots,v^{(n)}_{\hat{M}_j}$ in computing new scores for all hypotheses. The new LLM scores at step $j+1$ can be obtained with
\begin{align}
    S_{LM}(h_{j+1}^{(n)}) \approx & \log P_{LM}(v_1^{(n)},...,v_{\hat{M}_j}^{(n)}) \nonumber \\
    & + \log P_{LM}(v_{\hat{M}_j+1}^{(n)},...,v^{(n)}_{M_{j+1}^{(n)}}|v_1^{(n)},...,v_{\hat{M}_j}^{(n)}). \nonumber
\end{align}

\section{Experiments}
\label{sec:experiments}
\subsection{Conditions}
\label{ssec:conditions}
We conducted several experiments on the LibriHeavy corpus~\cite{kang2024libriheavy}, which includes 50k hours of English audio books.
We trained a CTC-AED model~\cite{watanabe2017hybrid,zhang2021wenet} using all three training subsets, i.e., small, medium, and large subsets, and also trained an in-domain NLM using formatted transcripts including casing and punctuation.
The CTC-AED model had an encoder network with a Conv2D module followed by 12 Conformer blocks, a decoder network with 3 unidirectional Transformer blocks, and a CTC output layer. 
In the Conv2D module, 80-dimensional Mel-filter bank features obtained every 10 msec were down-sampled by a factor of 6. 
We employed multi-head attention of 8 heads with 512 dimensions in total. The feed-forward network had one hidden layer of 2,048 units with ReLU activations. 
The in-domain NLM had 9 Transformer blocks with 256 dimensions and a shared embedding layer for input and output tokens.
The vocabulary size was 6K for both CTC-AED model and NLM, corresponding to the set of word pieces obtained from the LibriHeavy transcripts using the SentencePiece tokenizer~\cite{kudo2018sentencepiece}.
The number of parameters of the CTC-AED model and the NLM were 101M and 9M, respectively.
We applied SpecAugment in model training but did not use speed perturbation as in \cite{kang2024libriheavy}.

We also employed three public domain LLMs: OpenLLaMA 3B v2, OpenLLaMA 7B v2~\cite{openlm2023openllama}, and Mistral 7B v0.1~\cite{jiang2023mistral}, where the vocabulary size was 32K.
We evaluated the proposed method in two decoding modes, CTC prefix beam search~\cite{hannun2014first} and joint CTC-attention decoding~\cite{hori2017joint, watanabe2017hybrid}, where the former is frame-synchronous decoding and the latter is label-synchronous decoding. Although the frame-synchronous decoding is streamable, we used full utterance context in encoding to simplify the experiments.
We used language model weights in shallow and delayed fusion, which were tuned on the LibriHeavy dev set for each LM.
Evaluation metrics are word error rate (WER) and real-time factor (RTF). %
Decoding time was measured on an Intel Xeon (Skylake IBRS) CPU @ 2.4GHz with an NVIDIA V100 GPU.

\subsection{Results}
\label{ssec:results}
Table \ref{table:wer-and-rtf} shows WERs and RTFs for LibriHeavy test-clean (``lh-clean'') and test-other (``lh-other'') sets in different decoding conditions, where we compare the decoding modes, shallow fusion (SF) and delayed fusion (DF), and in-domain and large LMs, setting the beam size to 10.
To measure the RTF, we used only the first 200 utterances in ``lh-other''.
CTC prefix beam search without LM fusion is the baseline, which has the best RTFs, but high WERs. The WERs can be reduced by shallow fusion with the in-domain NLM, but RTF nonetheless increases significantly, even with the NLM probabilities computed on GPU. This is mainly due to frequent NLM calls during frame-synchronous decoding. 
We then evaluated delayed fusion using the shortest hypothesis approach described earlier.
With the in-domain NLM, with no tokenization mismatch, delayed fusion shows comparable WERs with shallow fusion, while reducing RTF from 0.082 to 0.066.
For the three LLMs, with the tokenization mismatch handled during decoding, delayed fusion produces better WERs than the baseline or NLM shallow fusion.
Although the RTF increases for the LLMs to 0.115, 0.141, and 0.142, these are still acceptable for real-time decoding.
With CTC-attention decoding, we see substantial improvements thanks to LLM fusion, although it does require more computation due to the use of the attention decoder and label-synchronous CTC. Note that, in label-synchronous decoding, delayed fusion based on the shortest hypothesis does not outperform shallow fusion in decoding speed (see SF vs. DF w/ NLM) because it does not reduce the number of LM calls.
However, delayed fusion still has the benefit of handling mismatched tokenization.  
Since the performance gap between LLMs is small, we report only the results with OpenLLaMA 3B model in the following.
\begin{table}[t]
\caption{Word error rate vs.\ real-time factor on LibriHeavy benchmarks\protect\footnotemark[1]. Delayed fusion (DF) was used for the LLMs, while shallow fusion (SF) was used for the in-domain NLM.}
\centering 
\label{table:wer-and-rtf}
\resizebox{.99\linewidth}{!}{
\setlength{\tabcolsep}{4pt}
\begin{tabular}{llccc}
\toprule
Decoding    & LM fusion & lh-clean & lh-other & RTF \\
\midrule
            & -                  &  3.47  &  6.41 & 0.018 \\
CTC prefix  & SF w/ NLM 9M       &  3.15  &  5.95 & 0.082 \\
beam        & DF w/ NLM 9M       &  3.16  &  5.93 & 0.066 \\
search      & DF w/ OpenLLaMA 3B &  3.05  &  5.68 & 0.115 \\
            & DF w/ OpenLLaMA 7B &  {\bf 3.01} &  5.67 & 0.141  \\
            & DF w/ Mistral 7B   &  3.02  & {\bf 5.63} & 0.142 \\
\midrule
            & -                  &  2.98  & 5.64 & 0.051 \\
CTC-        & SF w/ NLM 9M       &  2.94  & 5.61 & 0.091 \\
attention   & DF w/ NLM 9M       &  2.96  & 5.59 & 0.099 \\
decoding    & DF w/ OpenLLaMA 3B &  2.86  & 5.35 & 0.146 \\
            & DF w/ OpenLLaMA 7B &  2.84  & 5.29 & 0.170 \\
            & DF w/ Mistral 7B   &  {\bf 2.80} & {\bf 5.22} & 0.169 \\
\bottomrule
\end{tabular}
}
\end{table}
\footnotetext[1]{The WERs presented in this paper cannot strictly be compared with those in \cite{kang2024libriheavy} since our CTC-AED model is not compatible with their model in terms of vocabulary size, decoder architecture, and data augmentation.}

Table \ref{table:fixed_vs_delayed} compares delayed fusion strategies, the shortest-~hypothesis and fixed-interval methods, with N-best rescoring. Both strategies achieve better WERs than the baseline and N-best rescoring. Moreover, by changing the interval $I$, we can choose a suitable operating point considering the WER-RTF trade-off.
We also evaluate the combination of LLM delayed fusion with NLM shallow fusion, where the LM weight was evenly distributed across the two LMs during decoding, but the final hypothesis was selected with only the E2E and LLM scores (in line 13 of Algorithm \ref{algo:delayed_fusion}). 
The combined fusion effectively reduces the pruning error due to the delay, although it requires a certain overhead for the NLM.
\begin{table}[t]
\caption{Comparison of delayed fusion approaches and N-best rescoring in CTC prefix beam search. For fixed-interval fusion, we tested intervals of $I=16, 32$, and $64$, which correspond to 0.96s, 1.92s, and 3.84s, respectively, since one encoder frame is 60 ms.} 
\centering 
\label{table:fixed_vs_delayed}
\resizebox{.99\linewidth}{!}{
\setlength{\tabcolsep}{4pt}
\begin{tabular}{lcccc}
\toprule
                     & SF w/ NLM  & lh-clean & lh-other & RTF \\
\midrule
Baseline             &             &  3.47  &  6.41 & 0.018  \\
N-best resc. (N=10)  &             &  3.19  &  6.03 & 0.029  \\
Fixed-int. DF (I=64) &             &  3.09  &  5.78 & 0.041  \\
Fixed-int. DF (I=32) &             &  3.09  &  5.73 & 0.047  \\
Fixed-int. DF (I=16) &             &  3.08  &  5.73 & 0.063  \\
Shortest-hyp.  DF    &             &  3.05  & 5.68  & 0.115  \\
\midrule
Baseline w/ NLM     &  \checkmark &  3.15  &  5.95 & 0.082 \\
N-best resc. (N=10)  &  \checkmark &  3.07  &  5.79 & 0.089 \\
Fixed-int. DF (I=64) &  \checkmark &  3.05  &  5.70 & 0.099 \\
Fixed-int. DF (I=32) &  \checkmark &  3.05  &  5.68 & 0.107 \\
Fixed-int. DF (I=16) &  \checkmark &  3.05  &  5.67 & 0.122 \\
Shortest-hyp.  DF    &  \checkmark &  3.04  & 5.65  & 0.174 \\
\bottomrule
\end{tabular}
}
\end{table}

\begin{figure}[htb]
\centering
\centerline{\includegraphics[width=8.0cm]{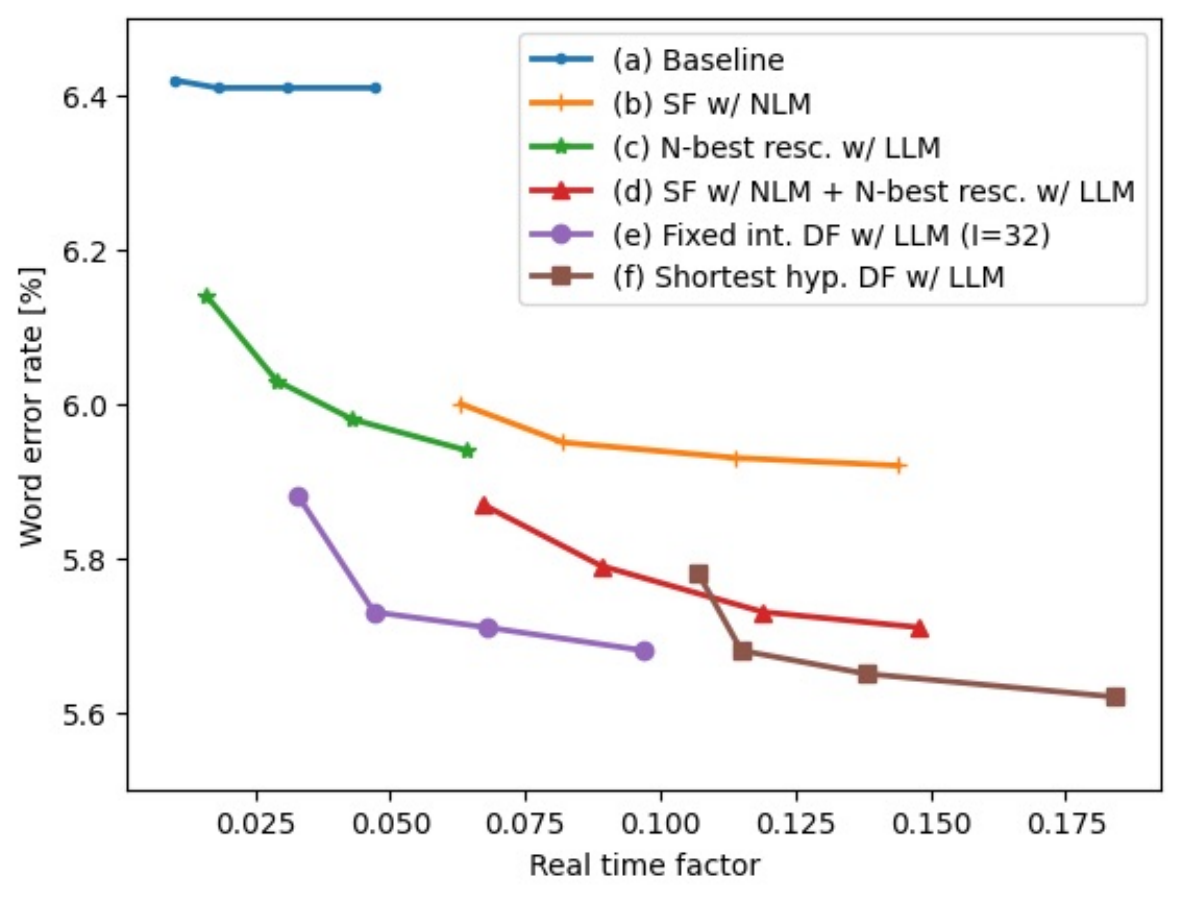}}
\caption{Delayed fusion vs. $N$-best rescoring for different beam sizes 5, 10, 15, and 20. WER and RTF were measured on ``lh-other''.}
\label{fig:different_beam_sizes}
\end{figure}

Figure \ref{fig:different_beam_sizes} compares delayed fusion and N-best rescoring performance for different beam sizes in CTC prefix beam search. The results indicate that delayed fusion, (e) \& (f), achieves lower WERs and RTFs than N-best rescoring, (c) \& (d).

Finally, Table \ref{table:comparison_with_llm_shallow_fusion} compares shallow and delayed fusion using the same LLM. For shallow LLM fusion, we trained another CTC-AED model from scratch using the LLM tokenizer, which had a 32K vocabulary. In CTC prefix beam search, delayed fusion (shortest hyp.) achieves a good accuracy comparable to or slightly better than shallow fusion, while delayed fusion is 2.2 times faster. In CTC-attention decoding, delayed fusion shows a comparable WER for ``lh-clean'' but a slightly worse WER for ``lh-other'', although it still has a certain speed benefit.
However, the advantage of delayed fusion is that it can avoid retraining of ASR models depending on the LLM.
\begin{table}[t]
\caption{Comparison with LLM shallow fusion, where LLM is OpenLLaMA 3B. For shallow LLM fusion, we trained another CTC-AED model with LLaMA tokenizer that vocabulary size was 32,000.}
\centering 
\label{table:comparison_with_llm_shallow_fusion}
\resizebox{.99\linewidth}{!}{
\setlength{\tabcolsep}{4pt}
\begin{tabular}{lclccc}
\toprule
Decoding & ASR vocab. & LM fusion & lh-clean & lh-other & RTF \\
\midrule
CTC prefix  & 6K  & -          &  3.47  &  6.41 & 0.018 \\
beam        & 32K & -          &  3.61  &  6.63 & 0.027 \\
search      & 6K  & DF w/ LLM  &  {\bf 3.05}  &  {\bf 5.68} & 0.115 \\
            & 32K & SF w/ LLM  &  {\bf 3.05}  &  5.75 & 0.257 \\
\midrule
CTC-        & 6K  & -          &  2.98  &  5.64 & 0.051 \\
attention   & 32K & -          &  3.03  &  5.62 & 0.062 \\
decoding    & 6K  & DF w/ LLM  &  {\bf 2.86}  &  5.35 & 0.146 \\
            & 32K & SF w/ LLM  &  {\bf 2.86}  &  {\bf 5.19} & 0.169 \\
\bottomrule
\end{tabular}
}
\end{table}

In summary, LLM delayed fusion achieves 4 - 13\% WER reduction (WERR) from the baseline and 3 - 7\% WERR from NLM shallow fusion (Table \ref{table:wer-and-rtf}). 
Furthermore, it provides lower WERs than $N$-best rescoring and NLM shallow fusion for the same decoding time (Fig. \ref{fig:different_beam_sizes}).
However, NLM shallow fusion followed by $N$-best LLM rescoring (plot (d) in Fig. \ref{fig:different_beam_sizes}) is competitive to delayed fusion ((e) \& (f)), where the relative WER difference is around 1 - 3 \%.
This is a small improvement, but an important advantage of delayed fusion is that it can be used for streaming decoding.
In some applications, such as live captioning, lattice/N-best rescoring is not an option, or can have a negative impact on the user experience; delayed fusion is applicable to a wider range of ASR applications.

\section{Conclusions}
\label{sec:conclusions}
In this paper, we proposed \textit{delayed fusion}, which applies LLM scores to first-pass ASR hypotheses with a delay during decoding and allows us to easily use pre-trained LLMs in ASR tasks. This method can reduce not only the number of hypotheses scored by the LLM but also the number of LLM inference calls. We can re-tokenize ASR hypotheses during decoding to compute LLM scores if ASR model and LLM employ different tokenizations.
We conducted experiments on the LibriHeavy corpus, applying delayed fusion with three public domain LLMs.
We demonstrated that (1) Delayed LLM fusion is fast enough compared to standard neural language model (NLM) fusion and (2) Delayed LLM fusion provides lower WERs than N-best LLM rescoring and standard NLM fusion.

Future work will include extensive evaluation of delayed fusion using different datasets, different metrics, e.g., GPU memory consumption, and E2E architectures including RNN Transducers~\cite{graves2012sequence}.

\section{ACKNOWLEDGEMENTS}
We thank David Rybach, Roger Hsiao, Dogan Can, and Pawel Swietojanski for useful suggestions on this work.

\bibliographystyle{IEEEbib}
\bibliography{refs}

\end{document}